\documentclass[10pt,letterpaper]{article}
\usepackage{cogsci}
\usepackage{pslatex}
\usepackage{apacite}
\usepackage{graphicx}
\usepackage[small]{caption} 
\usepackage{soul} 
\usepackage{color}
\usepackage{xcolor,colortbl}
\definecolor{orange}{HTML}{FCBF64}
\definecolor{skin}{HTML}{E79E97}

\renewcommand\bibliographytypesize{\small}

\title{Human few-shot learning of compositional instructions}
\author{{\large \bf Brenden M. Lake$^{1,2}$, Tal Linzen$^3$, and Marco Baroni$^{2,4}$} \\
$^1$New York University, $^2$Facebook AI Research, $^3$John Hopkins University, $^4$ICREA}

\begin{document}
\maketitle

\begin{abstract}
People learn in fast and flexible ways that have not been emulated by machines. Once a person learns a new verb ``dax,'' he or she can effortlessly understand how to ``dax twice," ``walk and dax,'' or ``dax vigorously.'' There have been striking recent improvements in machine learning for natural language processing, yet the best algorithms require vast amounts of experience and struggle to generalize new concepts in compositional ways. To better understand these distinctively human abilities, we study the compositional skills of people through language-like instruction learning tasks. Our results show that people can learn and use novel functional concepts from very few examples (few-shot learning), successfully applying familiar functions to novel inputs. People can also compose concepts in complex ways that go beyond the provided demonstrations. Two additional experiments examined the assumptions and inductive biases that people make when solving these tasks, revealing three biases: mutual exclusivity, one-to-one mappings, and iconic concatenation. We discuss the implications for cognitive modeling and the potential for building machines with more human-like language learning capabilities.

\textbf{Keywords:} 
concept learning; compositionality; word learning; neural networks
\end{abstract}

People use their compositional skills to make critical generalizations in language, thought, and action. Once a person learns a new concept ``photobombing'', she or he immediately understands how to ``photobomb twice'', ``jump and photobomb'', or ``photobomb vigorously.'' This example illustrates systematic compositionality, the algebraic capacity to understand and produce an infinite number of utterances from known components \cite{Chomsky:1957,Montague:1970a,Fodor1975}. This ability is central to how people can learn from limited amounts of experience \cite{Lake2016}, and uncovering its computational basis is an important open challenge.

There have been dramatic advances in machine language capabilities, yet the best algorithms require tremendous amounts of training data and struggle with generalization. These advances have been largely driven by neural networks, a class of models that has been long criticized for lacking systematic compositionality \cite{Fodor:Pylyshyn:1988,Marcus1998,Fodor:Lepore:2002,Marcus2003,Calvo:Symons:2014}. Neural networks have developed substantially since these classic critiques, yet recent work evaluated contemporary neural networks and found they still fail tests of compositionality \cite{LakeBaroni2018,Bastings:etal:2018,Loula2018}. To evaluate compositional learning, \citeA{LakeBaroni2018} introduced the SCAN dataset for learning instructions such as ``walk twice and jump around right,'' which were built compositionally from a set of primitive instructions (e.g., ``run'' and ``walk''), modifiers (``twice'' or ``around right''), and conjunctions (``and'' or ``after''). The authors found that modern recurrent neural networks can learn how to ``run'' and to ``run twice'' when both of these instructions occur in the training phase, yet fail to generalize to the meaning of ``jump twice'' when ``jump'' but not ``jump twice'' is included in the training data.

\begin{figure}[t]
\centering
\includegraphics[width=3.2in]{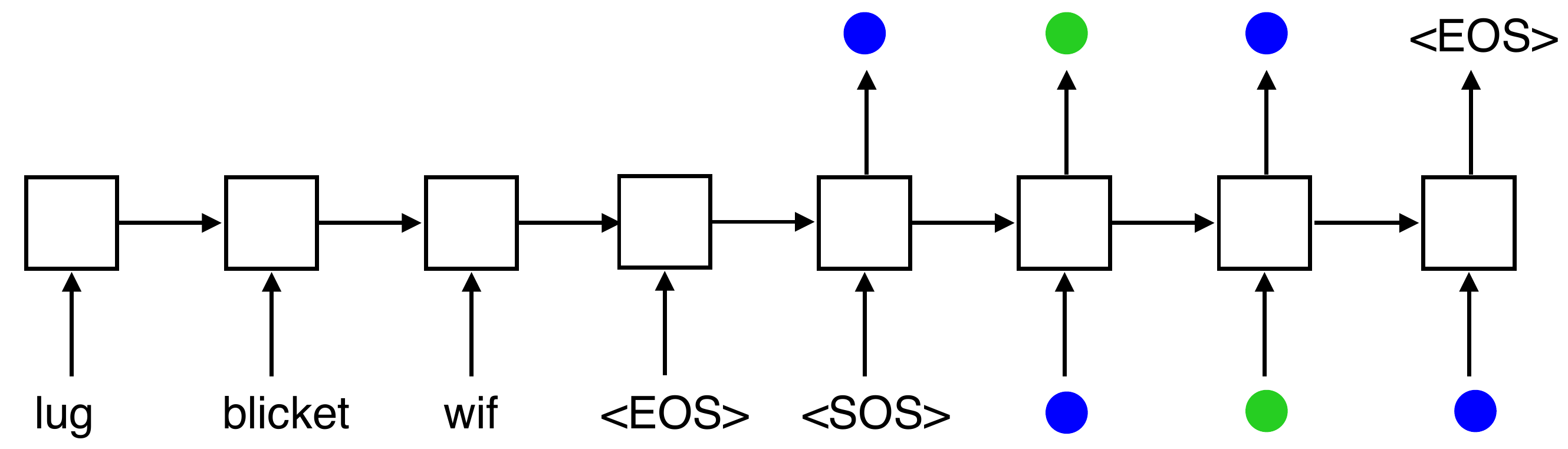}
\caption{A sequence-to-sequence (seq2seq) recurrent neural network applied to few-shot instruction learning. Instructions are provided in a novel language of pseudowords and processed with an encoder network (in this case, the instruction is ``lug blicket wif''), in order to generate an output sequence using a decoder network (``BLUE GREEN BLUE''). The symbols $<$EOS$>$ and $<$SOS$>$ denote end-of-sentence and start-of-sentence, respectively. The encoder (left) ends with the first $<$EOS$>$ symbol, and the decoder (right) begins with $<$SOS$>$.}
\label{fig_seq2seq}
\end{figure}

Classic arguments about the human ability to generalize have mostly
rested on thought experiments. The latter, however, might
underestimate facilitating factors, such as our knowledge of English,
on which we are undoubtedly relying when interpreting ``photobombing
twice''. In this paper, we study the scope and nature of people's
compositional learning abilities through artificial instruction
learning tasks that minimize reliance on knowledge of a specific
language. The tasks require mapping instructions to responses, where
an instruction is a sequence of pseudowords and a response is a
sequence of colored circles. These tasks follow the popular
sequence-to-sequence (seq2seq) framework and studied in
\citeA{LakeBaroni2018} and used to great effect in recent machine
learning \cite<e.g., machine translation;>{Sutskever2014}. 
Seq2seq tasks require a learner to first read a sequence of input symbols, 
and then produce a sequence of output symbols (Fig. \ref{fig_seq2seq}), whereby the input and output sequences can have different lengths.
This framework allows us to directly compare humans and recent
recurrent neural network architectures, while providing enough
flexibility and richness to study key aspects of compositional
learning. Moreover, the seq2seq problems investigated here present a
novel challenge for both human and machine learners: unlike standard seq2seq
benchmarks, which provide the learner with thousands of paired input and output examples, 
our ``few-shot learning" paradigm provides the learner with only a handful of training examples.

Our tasks differ from the artificial grammar learning
\cite{Reber:1967,Fitch:Friederici:2012}, rule learning
\cite{Marcus1999}, and program learning \cite{Stuhlmuller2010}
paradigms in that we do not ask participants to implicitly or
explicitly determine if items are grammatical. Instead, we ask them to
process input sequences in a pseudo-language in order to generate
output sequences (``meanings'').  Asking participants to associate new
words or sentences with visual referents is a standard practice in
psycholinguistics \cite<e.g.,>[and references
there]{Bloom:2000,Wonnacott:etal:2012}. Some of this work is
particularly close to ours in that it studies the biases underlying
linguistic generalization \cite<e.g.,>{HudsonKam:Newport:2009,Fedzechkina:etal:2016}. However, we are not aware of other studies that adopted the sequence-to-sequence language-to-meaning paradigm we are proposing here. Moreover, the biases studied in the earlier miniature language literature are more specific to grammatical phenomena attested in language (e.g., pertaining to linguistic syntax and morphology) than the basic generalization preferences we are exploring here. 

\begin{figure*}[t]
\centering
\includegraphics[width=7in]{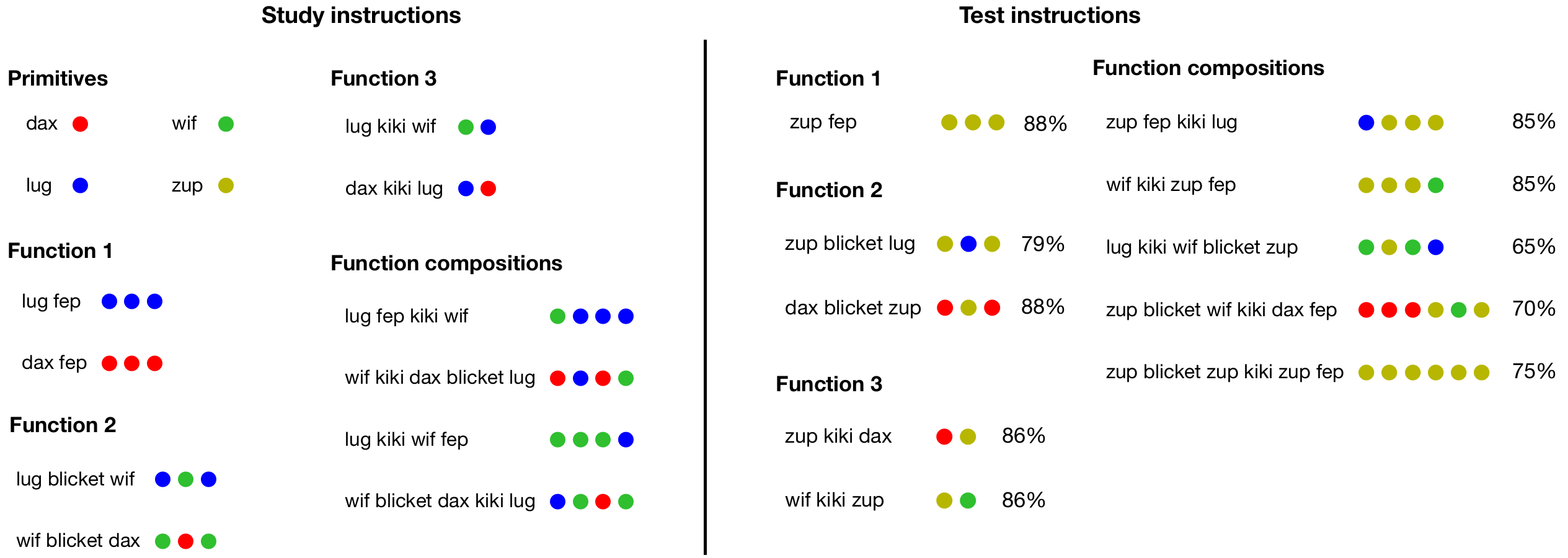}
\caption{Few-shot learning of instructions in Exp. 1. Participants learned to execute instructions in a novel language of pseudowords by producing sequences of colored circles. Generalization performance is shown next to each test instruction, as the percent correct across participants. The pseudowords and colors were randomized for each participant; the figure illustrates an example of such an assignment.}
\label{fig_miniscan}
\end{figure*}

\subsection{Experiment 1: Few-shot instruction learning}
Participants were asked to learn novel instructions from limited demonstrations. The task was inspired by the SCAN dataset for evaluating compositional learning in machines \cite{LakeBaroni2018}, adapted to be novel and tractable for human learners in the lab. Instead of following instructions in English, participants learned to interpret and execute instructions in a novel language of pseudowords (e.g., ``zup blicket lug'') by producing a sequence of abstract outputs (a sequence of colored circles; Fig.~\ref{fig_miniscan}). Some pseudowords were primitive instructions that correspond to a single output symbol, while other pseudowords are interpreted as functions that need to be applied to arguments to construct the output. As in SCAN, one primitive (``zup'') is only presented in isolation during study but is evaluated compositionally during test, appearing in each test instruction. To perform well, participants must learn the meaning of each function from just a small number of demonstrations, and then generalize to new primitives and more complex compositions than previously observed.

\subsubsection{Stimuli.}
The instructions consisted of seven possible pseudowords and the
output sequences consisted of four possible response symbols
(Fig.~\ref{fig_miniscan}). Four primitive pseudowords are direct
mappings from one input word to one output symbol (e.g., ``dax'' is
``RED'' and ``wif'' is ``GREEN''), and the other pseudowords are
functional terms that take arguments. To discourage a strategy based
on word-to-word translation into English, the functional terms could
not be easily expressed by single-word modifiers in English; they also formed
phrases whose order would be unnatural in English.

The meanings of the functions were as follows.
Function~1 (``fep'' in Fig.~\ref{fig_miniscan}) takes the preceding
primitive as an argument and repeats its output three times (``dax
fep'' is ``RED RED RED''). Function~2 (``blicket'') takes both the
preceding primitive and following primitive as arguments, producing
their outputs in a specific alternating sequence (``wif blicket dax''
is ``GREEN RED GREEN''). Last, Function~3 (``kiki'') takes both the
preceding and following strings as input, processes them, and
concatenates their outputs in reverse order (``dax kiki lug'' is
``BLUE RED''). We also tested Function~3 in cases where its arguments
were generated by the other functions, exploring function composition
(``wif blicket dax kiki lug'' is ``BLUE GREEN RED GREEN''). During the
study phase (see Methods below), participants saw examples that
disambiguated the order of function application for the tested
compositions (Function 3 takes scope over the other functions).

\subsubsection{Methods.}
Thirty participants in the United States were recruited using Amazon Mechanical Turk and the psiTurk platform \cite{Psiturk}. Participants were informed that the study investigated how people learn input-output associations, and that they would be asked to learn a set of commands and their corresponding outputs. Learning proceeded in a curriculum with four stages, with each stage featuring both a study phase and a test phase. In the first three stages, during the study phase participants learned individual functions from just two demonstrations each (Functions~1 through~3; Fig.~\ref{fig_miniscan}). In the final stage, participants learned to interpret complex instructions by combining these functions (Function compositions; Fig.~\ref{fig_miniscan}).

Each study phase presented participants with a set of example input-output mappings. For the first three stages, the study instructions always included the four primitives and two examples of the relevant function, presented together on the screen. For the last stage, the entire set of study instructions was provided together in order to probe composition. During the study phases, the output sequence for one of the study items was covered and participants were asked to reproduce it, given their memory and the other items on the screen. Corrective feedback was provided, and participants cycled through all non-primitive study items until all were produced correctly or three cycles were completed. The test phase asked participants to produce the outputs for novel instructions, with no feedback provided. The study items remained on the screen for reference, so that performance would reflect generalization in the absence of memory limitations. The study and test items always differed from one another by more than one primitive substitution (except in the Function 1 stage, where a single primitive was presented as novel argument to Function 1).  Some test items also required reasoning beyond substituting variables, and in particular understanding longer compositions of functions than were seen in the study phase.

The response interface had a pool of possible output symbols which could be clicked or dragged to the response array. The circles could be rearranged within the array or cleared with a reset button. The study and test set only used four output symbols, but the pool provided six possibilities (that is, there were two extra colors that were not associated to pseudowords), to discourage reasoning by exclusion. The assignment of nonsense words to colors and functions was randomized for each participant (drawn from nine possible nonsense words and six colors), and the first three stages were presented in random order.

We used several strategies to ensure that our participants were paying attention. First, before the experiment, participants practiced using the response interface and had to pass an instructions quiz; they cycled through the quiz until they passed it. Second, catch trials were included during the test phases, probing the study items rather than new items, with the answers clearly presented on the screen above. There was one catch trial per stage (except the last stage had two); a participants' test data was excluded if the participant missed two or more catch trials ($n=5$). Finally, test phases were also excluded if the corresponding study phases were not passed in the allotted time (13\% of remaining data).

\subsubsection{Recurrent neural networks.}
Standard sequence-to-sequence recurrent neural networks (RNNs; Fig.~\ref{fig_seq2seq}) failed to generalize from the study set to the test set. RNNs were trained using supervised learning on the 14 study instructions and evaluated on the test instructions (Fig.~\ref{fig_miniscan}), using the best overall architecture from \citeA{LakeBaroni2018} on the related SCAN benchmark (2-layer LSTM encoder and decoder, 200 hidden units per layer, a dropout probability of 0.5, no attention). This network (Fig.~\ref{fig_seq2seq}) consists of two neural networks working together: an encoder RNN that processes the instruction and embeds it as a vector, and a decoder RNN that decodes this vector as a sequence of output symbols. Another top architecture from \citeauthor{LakeBaroni2018} was also evaluated (1-layer LSTM encoder and decoder, 100 hidden units per layer, dropout 0.1, with attention). The training setup mimicked \citeauthor{LakeBaroni2018} but with 10,000 instruction presentations, corresponding to about 700 passes through the training data (epochs). Several variants of the architectures were also trained, repeatedly reducing the number of hidden units by half until there were only three hidden units per layer. Averaged across five random seeds, no architecture generalized better than 2.5\% correct on the test instructions, confirming \citeauthor{LakeBaroni2018}'s conclusion that seq2seq RNNs struggle with few-shot learning and systematic generalization.

\subsubsection{Results.}
Human participants showed an impressive ability to learn functions from limited experience and generalize to novel inputs, as summarized in Fig.~\ref{fig_miniscan}. In the first three stages, performance was measured separately for each functional term after exclusions through the above attention criteria. Average performance across participants was 84.3\% correct ($n=25$), counting sequences as correct only if every output symbol was correct. Measured for individual functions, accuracy was 88.0\% ($n = 25$) for Function~1, 83.3\% ($n=24$) for Function~2, and 86.4\% ($n = 22$) for Function~3.\footnote{The number of participants varies since data was included on the basis of passing the study phase. For comparison, the overall accuracy with no exclusions at all was 72.0\%.}

Participants were also able to compose functions together to interpret novel sequences of instructions. In the final stage, accuracy on complex instructions was 76.0\% ($n = 20$). People could generalize to longer and more complex instructions than previously observed, an ability that seq2seq neural networks particularly struggle with \cite{LakeBaroni2018}. During the study phase, the most complex instruction consisted of five input pseudowords requiring two function compositions, producing four output symbols. At test, most participants could successfully go beyond this, correctly processing six input pseudowords requiring three function compositions, producing six output symbols (72.5\% correct).

The pattern of errors showcases intriguing alternative hypotheses that
participants adopted. Some errors were suggestive of inductive biases
and assumptions that people bring to the learning task---principles
that are reasonable a priori and consistent with some but not all of the provided
demonstrations. For instance, many errors can be
characterized by a bias we term ``one-to-one,'' the
assumption that each input symbol corresponds to exactly one output
symbol, and that inputs can be translated one-by-one to outputs
without applying complex functional transformations. This
characterized 24.4\% of all errors.\footnote{These errors are defined as
  responses such that the input and output sequence
  have the same length, and each input primitive is replaced with its
  provided output symbol. Function words are replaced with an arbitrary output symbol.}
Other errors involved misapplication of
Function~3, which required concatenating its arguments in reverse
order. When participants made an error, they often concatenated but did not
reverse the argument (23.3\% of errors for instructions using Function~3), a bias we term ``iconic concatenation,'' referring to a preference for maintaining the order of the input symbols in the order of the output symbols. Forms of iconic concatenation are widely
attested in natural language, and constitute important biases in
language learning \cite{Haiman:1980,GoldinMeadow:etal:2008,deRuiter:etal:2018}.

In sum, people learn in several ways that go beyond powerful seq2seq neural networks. People can learn novel functions from as few as two examples and generalize in systematic ways, appropriately applying the functions to previously unused input variables. People can also compose these novel functions together in ways not observed during training. Finally, people appear to bring strong inductive biases to this learning challenge, which may contribute to both their learning successes and failures.

\begin{figure*}[t]
\centering
\includegraphics[width=6.5in]{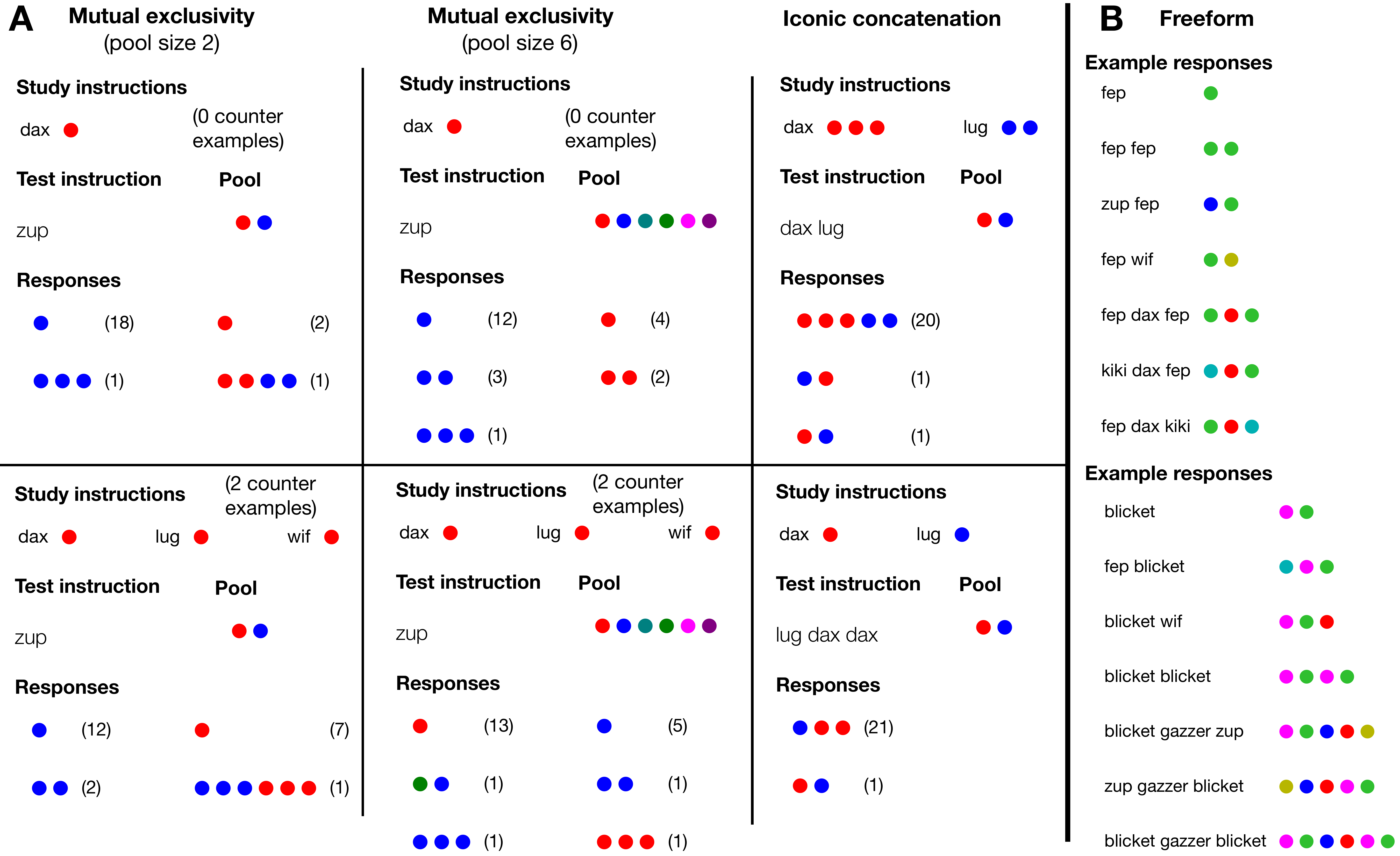}
\caption{Inductive biases in seq2seq word learning from Exp.~2 and 3. A: In Exp.~2, Participants were asked to respond to the Test instruction given the Study instructions, using only the symbols in the Pool. Shown are four examples trials (left and middle columns) examining mutual exclusivity with varying counter-evidence (varied across rows) and pool sizes (varied across columns), and two example trials (right column) examining iconic concatenation. All unique participant responses are shown with their frequency in parentheses. A canonical assignment of pseudowords and colors was used to aggregate the data, but it was randomized in the experiment. B: Responses from two participants in the Exp.~3 free-form task. The top participant was consistent with ME, one-to-one, and iconic concatenation, while the bottom participant was missing the one-to-one bias. For part B the words and colors are as-seen in the experiment.}
\label{fig_summary_bias}
\end{figure*}

\subsection{Experiment 2: Inductive biases in instruction learning}
This experiment investigated the inductive biases that appeared to influence the previous task. We devised a new set of seq2seq problems that were intentionally ambiguous and compatible with a number of possible generalizations, related to the ``poverty of the stimulus'' paradigm in experimental linguistics \cite{Wilson:2006,mccoy2018revisiting}. These problems provide a more direct window into people's inductive biases because the information provided is insufficient for deducing the correct answer. The design also parametrically varied the context under which the biases were evaluated to better understand their structure and scope.

This experiment studies the one-to-one and iconic concatenation biases
identified above, as well as the mutual exclusivity (ME) bias that has
been studied extensively in the developmental literature. Classic studies of ME present children with a familiar and an unfamiliar object (e.g., a ball and a spatula; \citeNP{Markman1988}), or two unfamiliar objects in which one is familiarized during the experiment \cite{Diesendruck2001}. When given the instruction ``show me the zup,'' children typically understand ``zup'' to refer to the novel object rather than acting as a
second name for the familiar object. In our instruction learning
paradigm, ME is operationalized as the inference that if ``dax'' means
``RED'', then ``zup'' is likely another response besides ``RED.''
Although Exp.~1 did not naturally lend itself to probing the effect of
the ME bias, we conjecture that it is because of the latter that
participants rapidly eliminated many degenerate solutions (such as all
strings referring to the same output item) in virtually any word
learning experiment. We thus want to study the impact of ME more
explicitly.

\subsubsection{Methods.}
Twenty-eight participants in the United States were recruited using Mechanical Turk and psiTurk. The instructions were as similar as possible to the previous experiment. In contrast, the curriculum of related stages in the previous experiment was replaced by 14 independent trials that evaluated biases under different circumstances. Each trial provided a set of study instructions (input-output mappings) and asked participants to make a judgment about a single new test instruction. To highlight the independence between trials, the pseudoword and colors were re-randomized for each trial from a larger set of 20 possible pseudowords and 8 colors. To emphasize the inductive nature of the task, participants were told that there were multiple reasonable answers for a given trial and were instructed to provide a reasonable guess.

\vspace{.2cm}
The trials were structured as follows. Six trials pertain to ME and whether participants are sensitive to counter-evidence and the number of options in the response pool (e.g., Fig~\ref{fig_summary_bias}A left and middle columns). Three trials pertain to iconic concatenation and how participants concatenate instructions together in the absence of demonstrations (e.g., Fig~\ref{fig_summary_bias}A right column). Three additional trials pertain to how people weigh ME versus one-to-one in judgments that necessarily violate one of these biases (not shown in figure). Finally, two catch trials queried a test instruction that was identical to a study instruction. The design minimized the risk that the biases could be learned from the stimuli themselves. None of the study instructions demonstrated how to concatenate, facilitating a pure evaluation of concatenation preferences. In the novel test trials, 6 instructions supported ME and 6 violated it, although both catch trials also supported ME. We did not explicitly control for the one-to-one bias. Missing a catch trial was the only criterion for exclusion ($n = 6$). There was no memory quiz for the study items since each contained just a few instructions.

\subsubsection{Results.}
There was strong evidence for each of the three inductive biases. The
classic mutual exclusivity (ME) effect was replicated within our
seq2seq learning paradigm. If ``dax'' means ``RED'', what is a
``zup''? As shown in the top-left cell of Fig~\ref{fig_summary_bias}A, most participants (18 of 22; 81.8\%) chose a single ``BLUE'' symbol as their response if the pool provided only ``RED'' and ``BLUE'' as options, and a larger fraction (20 of 22; 90.9\%) followed ME by choosing a (possibly multi-element) meaning different from ``RED.''

\begin{figure}[t]
\centering
\includegraphics[width=2.5in]{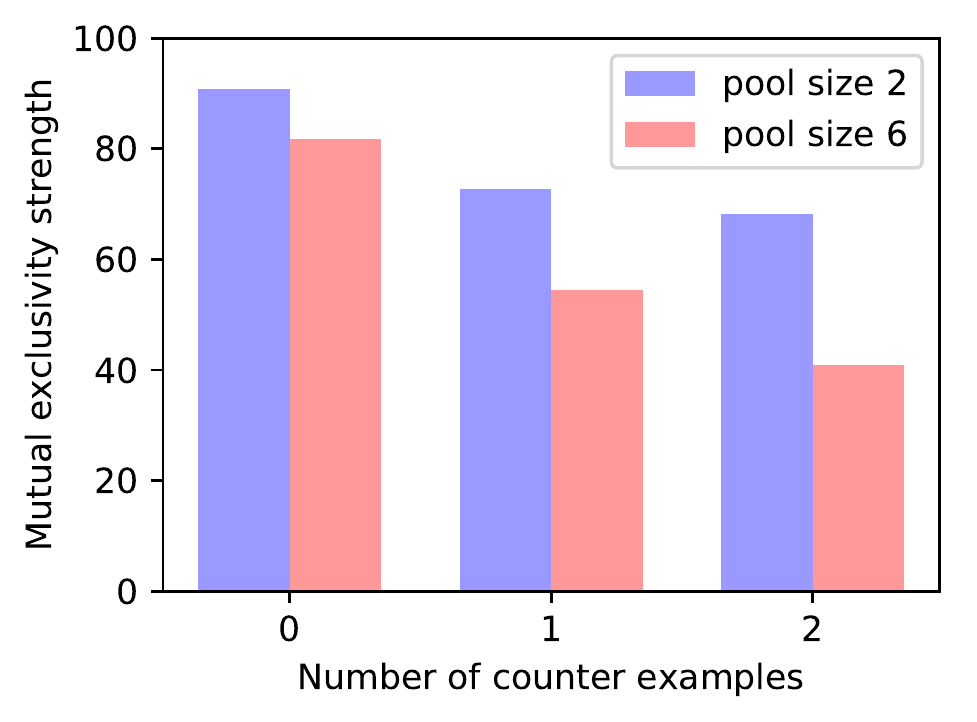}
\caption{The proportion of responses consistent with mutual exclusivity (y-axis) declines with the number of contradictory examples and the number of output symbols available in the response pool.}
\label{fig_summary_ME}
\end{figure}

While the ME effect was robust, it was sensitive to context and was not rigidly applied. The other ME trials examined the influence of two additional factors (Fig~\ref{fig_summary_bias}A left and middle columns): the number of contradictory examples provided (0--2; Fig~\ref{fig_summary_bias}A rows) and the number of output symbols available in the response pool (2~vs.~6; Fig~\ref{fig_summary_bias}A columns). With these two variables as fixed effects, we fit a logistic mixed model predicting whether or not a response was consistent with ME. Both the number of contradictory examples ($\beta = 1.76$, $SE=0.483$, $Z=3.64$, $p<0.001$) and pool size ($\beta=2.05$, $SE=0.698$, $Z=2.93$, $p<0.01$) were significant predictors, indicating that people were willing to override or weaken ME when faced with more ME counter-evidence (or equivalently in our case, positive evidence that ``RED'' is the right answer), or when more output symbols were available in the pool (Fig.~\ref{fig_summary_ME}). The second effect is intriguing. Although we leave a detailed analysis to future work, we conjecture that it stems from pragmatic reasoning on behalf of the participants: When five yet-to-be-named objects are in the pool, ME is such a weak heuristic that participants might conclude that the experiment is not asking them to rely on it.

There was strong confirmatory evidence for iconic concatenation. Across three trials that examined this bias in various forms, we found that 93.9\% ($n=22$) of responses were consistent with iconic concatenation, even though no examples of concatenation were provided during this experiment (Fig.~\ref{fig_summary_bias}A right column). In three trials where all of the output symbols in the pool were already assigned to unique pseudowords, participants had to choose between violating ME by reassigning an output symbol, or violating one-to-one by choosing a more complex  functional or multi-element meaning. Interestingly, the responses were evenly split (50.0\%) between following one principle versus the other.

Taken together, there was substantial support for three inductive biases in how people approach compositional learning in sequence-to-sequence mapping problem, confirming our hypotheses from Exp.~1. A drawback of this experiment's within-subjects design was the risk of judgments interfering with one another. The experiment used heavy randomization and mitigated the risk that the biases could be learned from the aggregate statistics of the stimuli, but these controls were not perfect. The next experiment addresses these concerns.

\subsection{Experiment 3: Inductive biases in free-form response}
In this experiment, participants responded to novel instructions without receiving any demonstrations, e.g., making plausible guesses for the outputs of instructions ``fep'', ``fep fep,'' and ``fep wif'' and how they relate to one another. This design offers the purest examination of people's assumptions since they have no relevant evidence about how to respond.

\subsubsection{Methods.}
Thirty participants in the United States were recruited using Mechanical Turk and psiTurk. The instructions were similar as possible to the previous experiments, using Exp.~2's wording emphasizing there are multiple reasonable answers and to provide a reasonable guess. Participants produced the output for seven novel instructions utilizing five possible pseudowords (Fig.~\ref{fig_summary_bias}B). Responses were entered on a single page, allowing participants to edit and maintain consistency. Participants also approved a summary view of their responses before submitting. There were six pool options, and the assignment of pseudowords and item order were random. One participant was excluded because she or he reported using an external aid in a post-test survey.

\subsubsection{Results.}
The results provide strong confirmatory evidence for the three key inductive biases: ME, iconic concatenation, and one-to-one. Although the task was highly under-determined, there was a substantial structure in the responses, unlike an untrained seq2seq recurrent neural network which would respond arbitrarily. The majority of participants (17 of 29; 58.6\%) responded in an analogous way to the participant shown at the top of Fig.~\ref{fig_summary_bias}B. This set of responses is perfectly consistent with all three inductive biases, assigning a unique output symbol to each input symbol and concatenating to preserve the input ordering. Other participants produced alternative hypotheses that followed some but not all the inductive biases. Overall, 23 of 29 participants (79.3\%) followed iconic concatenation, assigning consistent (but possibly multi-element) output sequences to individual input words (e.g., Fig.~\ref{fig_summary_bias}B bottom). In all but one of these cases, each input word was assigned a unique output sequence, abiding by mutual exclusivity (22 of 23; 95.7\%).

\subsection{Discussion and Conclusions}
People learn in fast and flexible ways not captured by today's algorithms. After learning how to ``dax'', people can immediately understand how to ``dax slowly'' or ``dax like you mean it.'' These types of inferences are critical to language learning and understanding, yet modern recurrent neural networks struggle to generalize in similarly systematic ways \cite{LakeBaroni2018,Loula2018}. To study these distinctively human abilities, we examined people's compositional skills in novel language-like instruction learning problems. The tasks followed the popular sequence-to-sequence (seq2seq) framework from machine learning, allowing humans and machines to be compared side-by-side. Experiment 1 examined how people learn novel instructions from examples, asking participants to interpret sequences of pseudowords by producing sequences of abstract output symbols. People could learn new functions from just two examples and successfully applied them to new inputs, while standard seq2seq recurrent neural networks (RNNs) failed to generalize. People could also handle longer sequences that require more compositions than previously observed, again surpassing the skills of powerful neural networks. Inspired by the errors participants made, Experiments 2 and 3 investigated inductive biases that constrain human learning, revealing that human learners draw upon mutual exclusivity (ME), iconic concatenation, and one-to-one in seq2seq word learning tasks.

More than a source of error, these biases provide important inductive constraints. If people interpreted the instruction as unanalyzable wholes, they would have no basis for generalization. Instead, people facilitate generalization by favoring hypotheses that assign unique and consistent meanings to individual words and follow certain input/output ordering constraints. As the final experiment shows, participants assume these characteristics before observing any data. The assumptions turn out to be powerful, characterizing most of the word meanings in Exp.~1 and the related SCAN benchmark, even though neither was designed with these biases in mind. Notably, the biases can mislead when learning function words; this was the case in many of the errors made in Exp.~1.

Future work should investigate the origin and scope of these biases through other compositional learning tasks. To the extent that our tasks evoke language learning, they could recruit biases known in the developmental literature such as mutual exclusivity \cite{Markman1988}. If the outputs are viewed as objects, one-to-one is related to the whole object assumption in word learning \cite{Macnamara:1982}. Alternatively, if the outputs are viewed as events or actions, iconic concatenation could be justified by aligning a description with its content in time \cite{deRuiter:etal:2018}. Another important line of future work should be providing a more explicit account of how the biases, which we observed emerging in participants' errors, are also aiding faster learning of the correct generalizations.

These insights from human learning could be fruitfully incorporated into machine learning. These biases could facilitate learning of seq2seq problems such as machine translation and semantic parsing, or related image2seq problems such as caption generation. Powerful seq2seq models do not have these inductive biases, suggesting a path to building more powerful and human-like learning architectures by incorporating them.

\subsection{Acknowledgments}
We thank the NYU ConCats group, Michael Frank, Kristina Gulordava, Germ\'{a}n Kruszewski, Roger Levy, and Adina Williams for helpful suggestions.
\newpage

\bibliographystyle{apacite}
\setlength{\bibleftmargin}{.125in}
\setlength{\bibindent}{-\bibleftmargin}
\bibliography{library_clean,marco}
\end{document}